\pdfoutput=1

\documentclass[11pt]{article}

\usepackage[]{emnlp2021}

\usepackage{times}
\usepackage{latexsym}

\usepackage[T1]{fontenc}

\usepackage[utf8]{inputenc}

\usepackage{microtype}

\usepackage[T1]{fontenc}

\usepackage[utf8]{inputenc}

\usepackage{microtype}
\usepackage{graphicx}
\usepackage{amsmath}

\usepackage[export]{adjustbox}
\usepackage{array}

\usepackage{enumitem}

\usepackage[ruled,vlined]{algorithm2e}
\DeclareMathAlphabet{\pazocal}{OMS}{zplm}{m}{n}

\usepackage{graphicx}               
\usepackage{tabularx}               
\newcolumntype{C}{>{\centering\arraybackslash}X}
\usepackage{multirow}               
\usepackage{diagbox}                
\usepackage{hhline}                 
\usepackage{color}                  
\usepackage{amssymb}                
\usepackage{mathtools}              
\usepackage{enumitem}               

\usepackage{makecell}
\usepackage{xcolor}
\usepackage{arydshln}
\usepackage{subcaption}
\usepackage{stix}
\usepackage{enumitem}

\definecolor{effectspancolor}{RGB}{0, 51, 125}
\definecolor{drugspancolor}{RGB}{44, 7, 110}

\usepackage{xparse}
\NewDocumentCommand{\heng}{ mO{} }{\textcolor{red}{\textsuperscript{\textit{Heng}}\textsf{\textbf{\small[#1]}}}}
\NewDocumentCommand{\tuan}{ mO{} }{\textcolor{blue}{\textsuperscript{\textit{Tuan}}\textsf{\textbf{\small[#1]}}}}
\NewDocumentCommand{\zixuan}{ mO{} }{\textcolor{orange}{\textsuperscript{\textit{Zixuan}}\textsf{\textbf{\small[#1]}}}}

\title{BERT might be Overkill: A Tiny but Effective Biomedical Entity Linker based on Residual Convolutional Neural Networks}

\author{Tuan Lai, Heng Ji, \textbf{ChengXiang Zhai}\\
	    Department of Computer Science\\
	    University of Illinois Urbana-Champaign\\
        \{tuanml2, hengji, czhai\}@illinois.edu
}

\begin{document}
\maketitle

\begin{abstract}



Biomedical entity linking is the task of linking entity mentions in a biomedical document to referent entities in a knowledge base. Recently, many BERT-based models have been introduced for the task. While these models have achieved competitive results on many datasets, they are computationally expensive and contain about 110M parameters. Little is known about the factors contributing to their impressive performance and whether the over-parameterization is needed. In this work, we shed some light on the inner working mechanisms of these large BERT-based models. Through a set of probing experiments, we have found that the entity linking performance only changes slightly when the input word order is shuffled or when the attention scope is limited to a fixed window size. From these observations, we propose an efficient convolutional neural network with residual connections for biomedical entity linking. Because of the sparse connectivity and weight sharing properties, our model has a small number of parameters and is highly efficient. On five public datasets, our model achieves comparable or even better linking accuracy than the state-of-the-art BERT-based models while having about 60 times fewer parameters. \footnote{The code is publicly available at \url{https://github.com/laituan245/rescnn_bioel}}
\end{abstract}

\section{Introduction}

Biomedical entity linking (EL)~\cite{zheng2014entity} is the task of linking biomedical mentions (e.g., diseases and drugs) to standard referent entities in a curated knowledge base (KB). For example, given the sentence  ``\textit{The average \textbf{\underline{NH3}} concentrations were low.}'', the mention \textit{\textbf{\underline{NH3}}} should be linked to the entity \texttt{KB:Ammonia}. Biomedical EL is an important research problem, with applications in many downstream tasks, such as biomedical question answering \cite{covidask}, information retrieval, and information extraction \cite{Wang2020COVID19LK,GEANet2020,lai2021joint,zhangetal2021fine}. In general, two main challenges of the EL task are: (1) \textit{ambiguity} - the same word or phrase can be used to refer to different entities; (2) \textit{variety} - the same entity can be referred to by different words or phrases. Unlike in the general domain, mentions in biomedical documents are relatively unambiguous \cite{dsouzang2015sieve,Li2017CNNbasedRF}. Building a system for biomedical EL involves primarily addressing the variety problem.

Recently, many BERT-based models have been introduced for biomedical EL \cite{Ji2020BERTbasedRF,biosyn,sapbert,liu2021learning}. While these models can achieve state-of-the-art results on many biomedical EL datasets, they are computationally expensive and contain about 110M parameters. Even though there are scientific labs that have a lot of computing resources, many researchers still have minimal access to large-scale computational power \cite{strubelletal2019energy}. Therefore, it is of practical importance to provide a more scalable solution for biomedical entity linking. Furthermore, the factors contributing to the success of these large BERT-based models remain unclear. And thus, it is not known whether the over-parameterization is needed to achieve competitive performance.


In this work, through a set of probing experiments, we shed some light on the inner workings of existing BERT models for biomedical EL. Surprisingly, the performance only changes slightly when the input word order is shuffled or when the attention scope is restricted. Based on these observations, we propose an effective convolutional neural network with residual connections (ResCNN) for the task. Because of the sparse connectivity and weight sharing properties, ResCNN has a small number of parameters and is highly efficient. Experiments on five datasets show that the performance of ResCNN is comparable to the state-of-the-art (SOTA) BERT-based models while having about 60 times fewer parameters.


\section{Methods}


In the following sections, we will first describe some preliminaries relating to the formulation of the EL problem and a general approach for the task (Sec. \ref{sec:preliminary}). We will then go into details about our probing experiments in Sec. \ref{sec:probing_experiments}. We will describe the design of our ResCNN model in Sec. \ref{sec:model_architecture}.

\begin{figure}[!t]
  \centering
  \includegraphics[width=0.925\linewidth]{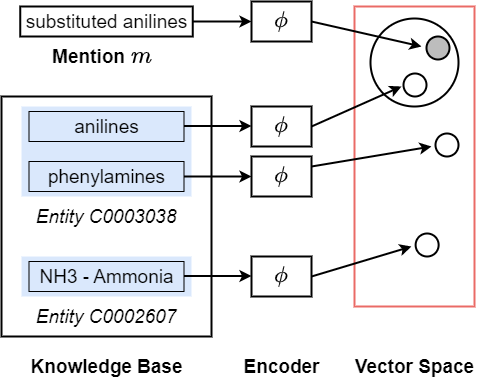}
  \caption{An illustration of the adopted approach to EL. In this example, the closest neighbor to the source  mention is the entity name \textit{anilines}. Therefore, this mention should be linked to the entity \textit{C0003038}.}
  \label{fig:dual_encoder_el}
\end{figure}

\subsection{Preliminaries} \label{sec:preliminary}

\paragraph{Problem Formulation} Given an entity mention $m$ from a biomedical text and a knowledge base (KB) consisting of $N$ entities $\mathcal{E} = \{e_1, e_2, ..., e_N\}$, the task is to find the entity $e_i \in \mathcal{E}$ that $m$ refers to. We assume that each entity is associated with a primary name and a list of alternative names. We denote the set of all names in the KB as $\mathcal{N} = \{n_1, n_2, ..., n_M\}$, where $M$ is the number of names. We use $T_m$ and $T_{n_j}$ to denote the textual forms of $m$ and $n_j$ respectively. Except for a list of names for each entity, we do not assume the availability of any other information in the KB (e.g., entity types or description sentences). Our formulation is general and suitable for a wide range of real-world settings.


\paragraph{General Approach} A general approach to EL is to train an encoder $\phi$ that encodes mentions and entity names into the same vector space \cite{gillicketal2019learning} (Figure \ref{fig:dual_encoder_el}). Before inference, we use $\phi$ to pre-compute embeddings for all the entity names in the KB. During inference, mentions are also encoded by $\phi$ and entities are retrieved using a simple distance function such as cosine similarity. In this work, we adopt this general approach, because it is more efficient and simpler than the two-stage retrieval and re-ranking systems \cite{wuetal2020scalable}. Several recent SOTA methods for biomedical EL also follow this approach. For example, \newcite{sapbert} models $\phi$ using SAPBERT, a BERT model pretrained on UMLS synonyms:
\begin{equation}\label{eq:baseline}
\begin{split}
    \phi(m) &= \text{SAPBERT}_{\text{CLS}}(T_m) \\
    \phi(n_j) &= \text{SAPBERT}_{\text{CLS}}(T_{n_j}) \; \forall\,n_j \in \mathcal{N}
\end{split}
\end{equation}
where $\text{SAPBERT}_{\text{CLS}}$ returns the final hidden state corresponding to the [CLS] token. Since SAPBERT was pre-trained on almost 12M pairs of synonyms, it can be directly used without further fine-tuning on the target task's training data. However, for several datasets, the performance can still be improved by training with task-specific supervision.




\begin{figure}[!t]
  \centering
  \includegraphics[width=0.925\linewidth]{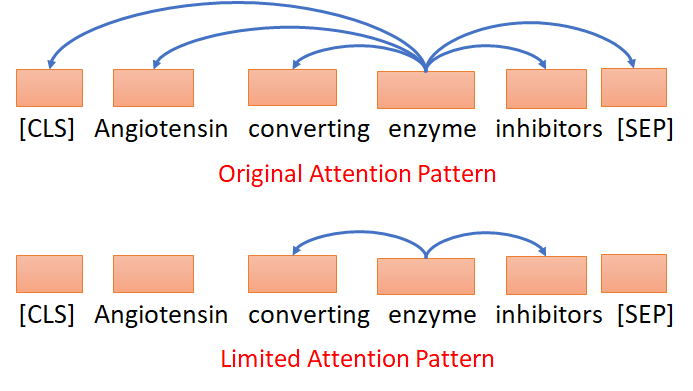}
  \caption{Attention scope restriction. In this example, the window size of the limited attention head is 3.}
  \label{fig:attention_scope_restriction}
\end{figure}

\subsection{Probing Experiments} \label{sec:probing_experiments}

Previous studies have shown that BERT can encode a wide range of syntactic and semantic features \cite{tenneyetal2019bert,jawaharetal2019bert}. However, it is unknown to what extent existing BERT models for biomedical EL utilize such rich linguistic signals. We take the first step towards answering this question by investigating the most basic aspects.


\paragraph{Word Order Permutation} We analyze whether BERT models fine-tuned for biomedical EL even consider one of the most fundamental properties of a sequence - the word order. In this probing experiment, we first train an EL model on the original (unshuffled) training set of a dataset. We then evaluate the model on the development set under the condition that the tokens of each mention/entity-name are shuffled.



\paragraph{Attention Scope Restriction} The self-attention mechanism of BERT makes each token in the input directly interact with every other token \cite{Vaswani2017AttentionIA}. As a result, the attention operation is quadratic to the input length. To analyze whether direct connections between distant tokens are crucial for biomedical EL, we conduct experiments where we restrict the attention scope to a local window (Figure \ref{fig:attention_scope_restriction}). We first train a BERT-based EL model on the provided training set of a dataset. We use the original attention mechanism during training. During evaluation, we limit the attention scope to a fixed window size by applying a masking operation:
\begin{equation}
\begin{split}
    \textbf{M}[i, j] &= \begin{cases} \mbox{1,} & \mbox{if } |i - j| \leq \lfloor w/2 \rfloor  \\ \mbox{-$\infty$,} & \mbox{otherwise} \end{cases}\\
    \text{Attention(\textbf{Q}, \textbf{K}, \textbf{V})} &= \text{softmax}\bigg(\frac{\textbf{M} \odot \textbf{Q}\textbf{K}^T}{\sqrt{p}}\bigg) \textbf{V} 
\end{split}
\end{equation}
where $\textbf{Q}$, $\textbf{K}$, and $\textbf{V}$ are the matrices of the queries, keys, and values (respectively) of an attention head \cite{Vaswani2017AttentionIA}. $w$ denotes the window size ($w$ is odd), $\odot$ denotes element-wise multiplication, and $p$ is a scaling factor. We restrict the attention scope of every token at every layer except for the [CLS] token at the last layer. We let the token attend to every other token at the last layer.




\renewcommand{\arraystretch}{1.1}
\begin{table*}[!ht]
\centering
\resizebox{\textwidth}{!}{%
\begin{tabular}{lcccccc}
\hline
\multicolumn{1}{c}{\multirow{2}{*}{Models}}               & \multicolumn{5}{c}{Top-1 Accuracy (on development sets)}                                                & \multirow{2}{*}{Avg. \% change} \\ \cline{2-6}
\multicolumn{1}{c}{}                                      & NCBI-d & BC5CDR-d & BC5CDR-c & MedMentions                         & COMETA &                                 \\ \hline

SAPBERT (Fine-Tuned) \shortcite{sapbert} & 91.1 & 90.9 & 98.2 & 54.4 & 74.9  &    \\
 \hline
Word Order Permutation & & & & \\
$\mdblkdiamond$ Shuffle unigrams & 88.2 & 90.2 & 94.0 & 53.2 & 65.6 & -4.58\% \\
$\mdblkdiamond$ Shuffle bigrams & 89.1 & 90.8 & 96.4 & 53.8 & 71.9 & -1.87\% \\
$\mdblkdiamond$ Shuffle trigrams & 90.5 & 91.0 & 97.7 & 54.0 & 73.1 & -0.87\% \\\hline
Attention Scope Restriction & & & & \\
$\mdblksquare$ Context size $= 3$ & 91.1 & 90.3 & 97.9 & 53.2 & 71.9 & -1.44\% \\
$\mdblksquare$ Context size $= 5$ & 91.2 & 90.9 & 97.6 & 53.8 & 73.4 & -0.74\% \\\hline
\end{tabular}%
}
\caption{Results of our conducted probing experiments with SAPBERT \cite{sapbert}.}
\label{tab:probing_results_sapbert}
\end{table*}
\begin{table*}[!ht]
\centering
\resizebox{0.8\textwidth}{!}{%
\begin{tabular}{lcccc}
\hline
\multicolumn{1}{c}{\multirow{2}{*}{Models}} & \multicolumn{3}{c}{Top-1 Accuracy (on test sets)} & \multirow{2}{*}{Avg. \% change} \\ \cline{2-4}
\multicolumn{1}{c}{} & NCBI-d & BC5CDR-d & BC5CDR-c &  \\ \hline
BIOSYN (Dense) \cite{biosyn} & 90.7 & 92.9 & 96.6 &  \\ \hline
Word Order Permutation & & & & \\
$\mdblkdiamond$ Shuffle unigrams & 67.0 & 77.0 & 74.8 & -21.94\% \\
$\mdblkdiamond$ Shuffle bigrams & 77.7 & 87.2 & 85.6 &  -10.62\% \\
$\mdblkdiamond$ Shuffle trigrams & 82.7 & 91.4 & 92.2 & -5.0\% \\ \hline
Attention Scope Restriction & & & & \\
$\mdblksquare$ Context size $= 3$ & 81.0 & 84.5 & 96.5 & -6.61\% \\ 
$\mdblksquare$ Context size $= 5$ & 78.8 & 87.5 & 96.5 & -6.35\% \\ \hline
\end{tabular}%
}
\caption{Results of our conducted probing experiments with BIOSYN \cite{biosyn}.}
\label{tab:probing_results_biosyn}
\end{table*}

\subsection{ResCNN for Biomedical Entity Linking} \label{sec:model_architecture}

As to be discussed in Section \ref{section:exps}, the performance of existing BERT models only changes slightly when the input word order is shuffled or when the attention scope is limited. These observations suggest that a simpler model that mainly focuses on capturing local interactions may perform as well as SOTA BERT-based models. A natural candidate that exhibits the desired properties is the convolutional neural network (CNN) architecture. CNNs have been empirically shown to be quite effective in capturing local features \cite{kim2014convolutional}. Furthermore, CNNs typically use fewer parameters than Transformer-based models because of their sparse connectivity and weight sharing properties. To this end, we introduce a simple but effective CNN with residual connections (ResCNN) for biomedical EL. Given an input text (e.g., a query mention or an entity name), ResCNN computes a vector representation for the input through several layers. 

\paragraph{Token Embedding Layer} We first use the BERT WordPiece tokenizer \cite{Wu2016GooglesNM} to split the original input text into a sequence of tokens. We then transform each token into an initial vector representation by re-using the first embedding layer of PubMedBERT \cite{Gu2020DomainSpecificLM}. This operation is very similar to using traditional word embeddings such as GloVe \cite{Pennington2014GloveGV}, and so it can be carried out efficiently. We keep the embedding layer fixed and do not tune its parameters during training. An advantage of WordPiece tokenization is that a relatively small vocabulary (e.g., 30,000 wordpieces) is sufficient to model large, naturally-occurring corpora. In contrast, the vocabulary size of traditional word embeddings is typically much larger.

\begin{figure}[!t]
  \centering
  \includegraphics[width=0.4\linewidth]{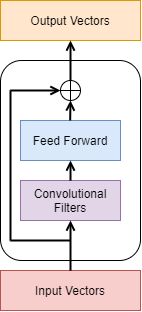}
  \caption{Encoding block of ResCNN.}
  \label{fig:rescnn_block}
\end{figure}

\paragraph{Encoding Layer} Our encoding layer consists of several \textit{encoding blocks} (Figure \ref{fig:rescnn_block}). Each block has multiple convolutional filters of varying window sizes \cite{kim2014convolutional}. Each filter is followed by an ReLU activation. We also employ a position-wise fully connected feed-forward network after applying the convolutional filters. In addition, there is a residual connection between the input and output of each encoding block. Residual connections alleviate the vanishing gradient problem \cite{He2016DeepRL}. Overall, our encoding blocks are quite similar to the Transformer encoder layers \cite{Vaswani2017AttentionIA}. However, we use local convolutional filters for feature extraction instead of the global attention mechanism.

\paragraph{Pooling Layer} To obtain the final vector representation for the input, we apply a pooling operation. In this work, we experiment with two different pooling strategies: (1) \textit{Max Pooling} \cite{kim2014convolutional} (2) \textit{Self-Attention Pooling} \cite{Zhu2018SelfAttentiveSE}.

We acknowledge that most of the components of our model are not novel as CNNs with residual links have been used in other tasks \cite{conneauetal2017deep,huangwang2017deep}. Nevertheless, our work provides evidence for the importance of carefully justifying the complexity of existing or newly proposed models. Depending on the specific task, a lightweight model may perform as well as the large BERT-based models. Also, our proposed ResCNN achieves SOTA performance on several datasets while being even more efficient than previous CNN-based or RNN-based methods (Sec. \ref{section:exps}).
\section{Experiments} \label{section:exps}

\paragraph{Data and Experimental Setup} We experiment across five different datasets: NCBI \cite{NCBIdataset}, BC5CDR-c and BC5CDR-d \cite{Li2016BioCreativeVC}, MedMentions \cite{Mohan2019MedMentionsAL}, and COMETA \cite{cometadataset}. For each dataset, we follow the data split by \citet{sapbert}. It is worth highlighting that even though  the five datasets can all be categorized as ``biomedical datasets'', they have very different characteristics. For example, while MedMentions was constructed by annotating scientific papers, COMETA was built by crawling Reddit (a social media forum). We report results in terms of top-1 accuracy.  Details about the hyperparameters are in the appendix.

\paragraph{Probing Results (SAPBERT)} Table \ref{tab:probing_results_sapbert} shows the results of our probing experiments with SAPBERT \cite{sapbert}. When the inputs' unigrams are randomly re-ordered, the performance of SAPBERT only drops by about 4.58\% on average. The difference is even less noticeable when we shuffle trigrams instead of unigrams. Therefore, SAPBERT is highly insensitive to word-order randomization. These results agree with recent studies on general-domain BERT models \cite{Pham2020OutOO,sinha2021masked}. Table \ref{tab:probing_results_sapbert} also shows that the performance of SAPBERT only changes slightly when the attention scope is limited.

\renewcommand{\arraystretch}{1.1}
\begin{table*}[!ht]
\centering
\resizebox{\textwidth}{!}{%
\begin{tabular}{lcccccc}
\hline
\multicolumn{1}{c}{\multirow{2}{*}{Models}}               & \multicolumn{5}{c}{Top-1 Accuracy (on test sets)}                                                & \multirow{2}{*}{Nb. Parameters} \\ \cline{2-6}
\multicolumn{1}{c}{}                                      & NCBI-d & BC5CDR-d & BC5CDR-c & MedMentions                         & COMETA &                                 \\ \hline
BNE \shortcite{Phan2019RobustRL}         & 87.7   & 90.6     & 95.8     & -                              & -           &  4.1M \\
CNN-based Ranking \shortcite{Chen2020ALN} & 89.6 & - & - & - & - & 4.6M
\\\hline
SAPBERT (Fine-Tuned) * \shortcite{sapbert} & 92.3   & 93.2     & 96.5     & 50.4                           & 75.1  &   110M                              \\
BIOSYN * \shortcite{biosyn}                & 91.1   & 93.2     & 96.6     & OOM                           & 71.3       &   110M  \\
BIOSYN (init. w/ SAPBERT) * & \textbf{92.5} & \textbf{93.6} & 96.8 & OOM & 77.0 & 110M \\
 \hline
ResCNN (Self-Attention Pooling) & 92.2 & 93.2 & \textbf{96.9} & \textbf{55.0} & 79.4 & \textbf{1.8M} \\
ResCNN (Max Pooling)  & 92.4 &  93.1 &  96.8 & 53.5 & \textbf{80.1} & \textbf{1.7M} \\\hline
\end{tabular}%
}
\caption{Overall test results on the five biomedical EL datasets. ``-'' denotes results not reported in the cited paper. The symbol * denotes BERT-based models. OOM stands for out-of-memory.}
\label{tab:overall_test_results}
\end{table*}

\renewcommand{\arraystretch}{1.1}
\begin{table*}[!ht]
\centering
\resizebox{\textwidth}{!}{%
\begin{tabular}{lcccccccccc}
\hline
\multicolumn{1}{c}{\multirow{2}{*}{Models}} & \multicolumn{2}{c}{NCBI-d} & \multicolumn{2}{c}{BC5CDR-d} & \multicolumn{2}{c}{BC5CDR-c} & \multicolumn{2}{c}{MedMentions} & \multicolumn{2}{c}{COMETA} \\ \cline{2-11} 
\multicolumn{1}{c}{} & \multicolumn{1}{l}{CPU} & GPU & \multicolumn{1}{l}{CPU} & GPU & \multicolumn{1}{l}{CPU} & GPU & \multicolumn{1}{l}{CPU} & GPU & \multicolumn{1}{l}{CPU} & GPU \\ \hline
SAPBERT \shortcite{sapbert} & 534s & 58s & 551s & 66s & 3478s & 276s & OOM & 6269s  & 6156s & 470s \\
ResCNN + Max Pooling & 33s & 18s & 35s & 21s & 169s & 69s & 3274s & 1565s & 289s & 109s  \\ \hdashline
Speedup (compared to SAPBERT) & 16.2x & 3.2x & 15.7x & 3.1x & 20.6x & 4.0x & - & 4.0x & 21.3x & 4.3x \\\hline
\end{tabular}%
}
\caption{Inference time of different models on CPU and GPU. OOM stands for out-of-memory.}
\label{tab:training_inference_time}
\vspace{-6mm}
\end{table*}

\paragraph{Probing Results (BIOSYN)} We have also experimented with BERT models trained on the BIOSYN framework \cite{biosyn}. We directly use the trained BERT models downloaded from \url{https://github.com/dmis-lab/BioSyn}. Table \ref{tab:probing_results_biosyn} shows the results of our conducted probing experiments with BIOSYN. Note that the authors of BIOSYN only provided the trained checkpoints for NCBI-d, BC5CDR-d, and BC5CDR-c. Overall, the changes are more prominent for models trained on BIOSYN than for SAPBERT. Nevertheless, the performance only drops by about 5.0\% on average when the inputs' trigrams are randomly re-ordered. The performance also only changes by 6.61\% when the attention window is set to be 3.

\paragraph{Entity Linking Accuracy} Table \ref{tab:overall_test_results} shows the linking performance of various models. Despite having less than 2M parameters, our CNN-based models achieve better results than the previous BERT-based SOTA systems on three of the datasets. It is worth noting that SAPBERT \cite{sapbert} was pre-trained on almost 12M pairs of UMLS synonyms. Without such pre-training, our lightweight models still match the performance of SAPBERT.

\paragraph{Inference Time} Table \ref{tab:training_inference_time} shows the speed of various models on CPU and on GPU. Compared to SAPBERT, our model is about 3 to 4 times faster on GPU and about 15 to 20 times faster on CPU. It takes less time to run our model on CPU than running SAPBERT on GPU. These results demonstrate the efficiency of our proposed model.
\section{Conclusions and Future Work}
Our work has shown that while BERT has been widely used for many NLP tasks, it is sometimes an overkill for some tasks, in which case, a simpler model can be as effective as BERT and is often much more efficient. An interesting future direction is to study further how to systematically simplify/compress BERT based on the insights obtained using probing experiments to increase efficiency while maintaining effectiveness. We plan to extend our work to other domains as well as other information extraction tasks \cite{lai2020joint,linetal2020joint,wen2021resin,lai2021context,li2020gaia}.


\section*{Acknowledgement}
This research is based upon work supported by the Molecule Maker Lab Institute: An AI Research Institutes program supported by NSF under Award No. 2019897, NSF No. 2034562, Agriculture and Food Research Initiative (AFRI) grant no. 2020-67021-32799/project accession no.1024178 from the USDA National Institute of Food and Agriculture, and U.S. DARPA KAIROS Program No. FA8750-19-2-1004. The views and conclusions contained herein are those of the authors and should not be interpreted as necessarily representing the official policies, either expressed or implied, of DARPA, or the U.S. Government. The U.S. Government is authorized to reproduce and distribute reprints for governmental purposes notwithstanding any copyright annotation therein.

\bibliography{anthology,custom}
\bibliographystyle{acl_natbib}

\appendix
\section{Reproducibility Checklist} \label{sec:reproducibility_checklist}
\renewcommand{\arraystretch}{1.1}
\begin{table*}[!ht]
\centering
\resizebox{\textwidth}{!}{%
\begin{tabular}{lcccccc}
\hline
\multicolumn{1}{c}{\multirow{2}{*}{Models}}               & \multicolumn{5}{c}{Top-1 Accuracy (on test sets)}                                                & \multirow{2}{*}{Nb. Parameters} \\ \cline{2-6}
\multicolumn{1}{c}{}                                      & NCBI-d & BC5CDR-d & BC5CDR-c & MedMentions                         & COMETA &                                 \\ \hline
ResCNN (Self-Attention Pooling) & 92.9 & 97.0 & 99.5 & 55.0 & 79.3 & 1.8M \\
ResCNN (Max Pooling)  & 95.0 & 91.8 & 99.3  & 53.8 & 79.9 & 1.7M \\\hline
\end{tabular}%
}
\caption{Final validation scores of our ResCNN models on the five biomedical EL datasets.}
\label{tab:rescnn_validation_scores}
\end{table*}

In this section, we present the reproducibility information of the paper.

\paragraph{Implementation Dependencies Libraries} Pytorch 1.6.0 \cite{Paszke2019PyTorchAI}, Transformers 4.4.2 \cite{wolfetal2020transformers}, Numpy 1.19.5 \cite{Harris2020ArrayPW}, CUDA 11.0.

\paragraph{Computing Infrastructure} The experiments were conducted on a server with Intel(R) Xeon(R) Gold 5120 CPU @ 2.20GHz and NVIDIA Tesla V100 GPUs. The allocated RAM is 191.9G. GPU memory is 16G.

\paragraph{Datasets} NCBI-d, BC5CDR-c, and BC5CDR-d can be downloaded from \url{https://github.com/dmis-lab/BioSyn}. MedMentions can be downloaded from \url{https://github.com/chanzuckerberg/MedMentions}. COMETA can be downloaded from \url{https://github.com/cambridgeltl/cometa}.

\paragraph{Average Runtime} We have presented the information of the inference time of our models in the main paper.

\paragraph{Number of Model Parameters} We have discussed about the models' sizes in the main paper.

\paragraph{Hyperparameters of Best-Performing Models} Each of our best ResCNN models consists of 4 encoding blocks. Each encoding block has 100 filters of kernel size 1, 100 filters of kernel size 3, and 100 filters of kernel size 5 (300 filters in total). The learning rate used for training our models is set to be 0.001. We use the Adam optimizer to train the ResCNN models. We use Huggingface’s Transformer library to experiment with different BERT models \cite{wolfetal2020transformers}.

\paragraph{Expected Validation Performance} For each of the MedMentions and COMETA datasets, we report the test performance of the checkpoint with the best validation score in the main paper. For each of the remaining three datasets, we use the corresponding development (dev) set to search for the hyperparameters and then train on the traindev (train+dev) set to report the final performance \cite{biosyn}. The final validation scores of our ResCNN models are shown in the Table \ref{tab:rescnn_validation_scores}.

\end{document}